\documentclass{article}

\PassOptionsToPackage{numbers, compress}{natbib}

\usepackage[preprint]{neurips_2024}

\usepackage[utf8]{inputenc} 
\usepackage[T1]{fontenc}    
\usepackage{hyperref}       
\usepackage{url}            
\usepackage{booktabs}       
\usepackage{amsfonts}       
\usepackage{nicefrac}       
\usepackage{microtype}      
\usepackage{xcolor}         

\usepackage{graphicx}
\usepackage[linesnumbered,ruled,vlined]{algorithm2e}
\usepackage{amsmath}
\usepackage{abbrv}
\usepackage{pgfplotstable}

\newcommand{\parag}[1]{{\noindent\bf{#1}}}

\title{DiffSF: Diffusion Models for Scene Flow Estimation}

\author{%
  Yushan Zhang \quad Bastian Wandt \quad Maria Magnusson \quad Michael Felsberg\\
  Linköping University\\
  \texttt{\{firstname.lastname\}@liu.se}\\
}

\begin{document}

\maketitle

\begin{abstract}
Scene flow estimation is an essential ingredient for a variety of real-world applications, especially for autonomous agents, such as self-driving cars and robots. 
While recent scene flow estimation approaches achieve reasonable accuracy, their applicability to real-world systems additionally benefits from a reliability measure. 
Aiming at improving accuracy while additionally providing an estimate for uncertainty, we propose \emph{DiffSF} that combines transformer-based scene flow estimation with denoising diffusion models. 
In the diffusion process, the ground truth scene flow vector field is gradually perturbed by adding Gaussian noise. 
In the reverse process, starting from randomly sampled Gaussian noise, the scene flow vector field prediction is recovered by conditioning on a source and a target point cloud. 
We show that the diffusion process greatly increases the robustness of predictions compared to prior approaches resulting in state-of-the-art performance on standard scene flow estimation benchmarks. 
Moreover, by sampling multiple times with different initial states, the denoising process predicts multiple hypotheses, which enables measuring the output uncertainty, allowing our approach to detect a majority of the inaccurate predictions. 
The code is available at \href{https://github.com/ZhangYushan3/DiffSF}{https://github.com/ZhangYushan3/DiffSF}.
\end{abstract}

\section{Introduction}

Scene flow estimation is an important research topic in computer vision with applications in various fields, such as autonomous driving~\cite{menze2015object} and robotics~\cite{seita2023toolflownet}. 
Given a source and a target point cloud, the objective is to estimate a scene flow vector field that maps each point in the source point cloud to the target point cloud. 
Many studies on scene flow estimation aim at enhancing accuracy and substantial progress has been made particularly on clean, synthetic datasets. 
However, real-world data contains additional challenges such as severe occlusion and noisy input, thus requiring a high level of robustness when constructing models for scene flow estimation.

\begin{figure}[t]
    \centering
    \includegraphics[width=.98\linewidth]{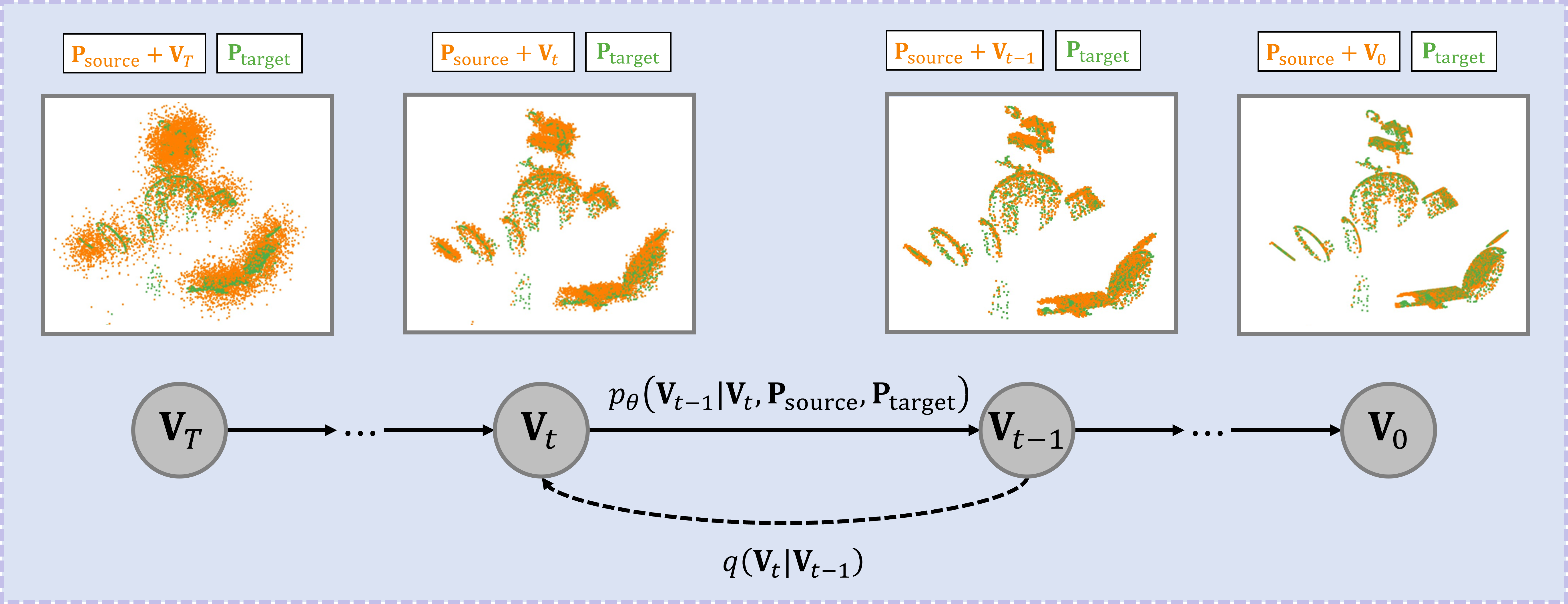}
    \caption{\textbf{Diffusion process.} In the forward process, we start from a ground truth scene flow vector field $\mathbf{V}_0$ and gradually add noise to it until we reach $\mathbf{V}_T$, which is completely Gaussian noise.
    In the reverse process, we recover the scene flow vector field $\mathbf{V}_0$ from the randomly sampled noisy vector field $\mathbf{V}_T$ conditioned on the source point cloud $\mathbf{P}_\mathrm{source}$ and the target point cloud $\mathbf{P}_\mathrm{target}$.}
    \label{Diffusion}
\end{figure}

Recently, Denoising Diffusion Probabilistic Models (DDPMs) have not only been widely explored in image generation~\cite{ho2020denoising, rombach2022high} but also in analysis tasks, e.g.\ detection~\cite{chen2023diffusiondet}, classification~\cite{han2022card}, segmentation~\cite{baranchuk2021label, gu2022diffusioninst}, optical flow~\cite{saxena2024surprising}, human pose estimation~\cite{holmquist2023diffpose}, point cloud registration~\cite{jiang2024se}, etc. 
Drawing inspiration from the recent successes of diffusion models in regression tasks and 
recognizing their potential compatibility with scene flow estimation, 
we formulate scene flow estimation as a diffusion process following DDPMs~\cite{ho2020denoising} as shown in Figure~\ref{Diffusion}. 
The forward process initiates from the ground truth scene flow vector field and gradually introduces noise to it. 
Conversely, the reverse process is conditioned on the source and the target point cloud 
and is tasked to reconstruct the scene flow vector field based on the current noisy input. 
To learn the denoising process, a new network is proposed inspired by state-of-the-art scene flow estimation methods FLOT~\cite{puy2020flot} and GMSF~\cite{zhang2024gmsf}. 

Previous methods~\cite{zhang2024gmsf, cheng2023multi, wang2022matters, cheng2022bi} usually suffer from inaccuracies when occlusions occur or when dealing with noisy inputs. 
During inference, based on the fixed parameters learned during training, 
they cannot provide information about their inaccurate predictions, 
which might lead to problems in safety-critical downstream tasks. 
Our proposed method approaches this problem in two aspects:
First, 
denoising diffusion models are capable of handling noisy data by modeling stochastic processes. 
The noise caused by sensors in the real world is filtered out, which allows the model to focus on learning underlying patterns. 
By learning feature representations that are robust to noise, the prediction accuracy is improved. 
Second, 
since the diffusion process introduces randomness into the inherently deterministic prediction task, it can provide a measure of uncertainty for each prediction by averaging over a set of hypotheses, notably without any modifications to the training process.
Extensive experiments on multiple benchmarks, FlyingThings3D~\cite{mayer2016large}, KITTI Scene Flow~\cite{menze2015object}, and Waymo-Open~\cite{sun2020scalability}, demonstrate state-of-the-art performance of our proposed method. 
Furthermore, we demonstrate that the predicted uncertainty correlates with the prediction error, establishing it as a reasonable measure that can be adjusted to the desired certainty level with a simple threshold value.

To summarize, our contributions are: 
(1) We introduce DiffSF, leveraging diffusion models to solve the full scene flow estimation problem, where the inherent noisy property of the diffusion process filters out noisy data, thus, increasing the focus on learning the relevant patterns. 
(2) DiffSF introduces randomness to the scene flow estimation task, which allows us to predict the uncertainty of the estimates without being explicitly trained for this purpose.
(3) We develop a novel architecture that combines transformers and diffusion models for the task of scene flow estimation, improving both accuracy and robustness for a variety of datasets. 

\section{Related Work}

\parag{Scene Flow Estimation}
has rapidly progressed since the introduction of FlyingThings3D~\cite{mayer2016large}, KITTI Scene Flow~\cite{menze2015object}, and Waymo-Open~\cite{sun2020scalability} benchmarks. Many existing methods~\cite{behl2017bounding,ma2019deep,menze2015object,ren2017cascaded,sommer2022sf2se3,vogel20153d,yang2021learning} assume scene objects are rigid and break down the estimation task into sub-tasks involving object detection or segmentation, followed by motion model fitting. While effective for autonomous driving scenes with static background and moving vehicles, these methods struggle with more complex scenes containing deformable objects, and their non-differentiable components impede end-to-end training without instance-level supervision.
Recent advancements in scene flow estimation focus on end-to-end trainable models and are categorized into encoder-decoder, coarse-to-fine, recurrent, soft correspondence methods, and runtime optimization-based methods. 
Encoder-decoder techniques, exemplified by FlowNet3D~\cite{liu2019flownet3d,wang2020flownet3d++} and HPLFlowNet~\cite{gu2019hplflownet}, utilize neural networks to learn scene flow by adopting an hourglass architecture. 
Coarse-to-fine methods, such as PointPWC-Net~\cite{wu2020pointpwc}, progressively estimate motion from coarse to fine scales, leveraging hierarchical feature extraction and warping. 
Recurrent methods like FlowStep3D~\cite{kittenplon2021flowstep3d}, PV-RAFT~\cite{wei2021pv}, and RAFT3D~\cite{teed2021raft} iteratively refine the estimated motion, thus enhancing accuracy. 
Some approaches like FLOT~\cite{puy2020flot}, STCN\cite{li2022sctn}, and GMSF~\cite{zhang2024gmsf} frame scene flow estimation as an optimal transport problem, employing convolutional layers and point transformer modules for correspondence computation. 
Different from the previously mentioned methods, which are fully trained and supervised offline, the runtime optimization-based methods~\cite{li2021neural,li2023fast,chodosh2023re} are optimized during the evaluation time based on each pair of inputs. 
While these methods have the advantage of without the need for training datasets, it also means that they can not take advantage of large-scale training datasets. Due to the online optimization, they also suffer from slow inference speed. Moreover, most of them focus only on autonomous driving scenes. 
On the other hand, we aim to estimate the scene flow of more general scenarios. 
Our proposed method takes the current state-of-the-art soft correspondence method GMSF~\cite{zhang2024gmsf} as a baseline. 
Given the fact that being able to indicate uncertainty of the estimation is an important feature for safety-critical downstream tasks, 
we propose to leverage the diffusion models for this purpose, whose ability of uncertainty indication has been proven by other relevant research areas~\cite {han2022card, saxena2024surprising}. 

\parag{Diffusion Models for Regression.}
Diffusion models have been widely exploited for image generation~\cite{ho2020denoising, rombach2022high}.
Beyond their capacity to generate realistic images and videos, researchers have also explored their potential to approach regression tasks. 
CARD~\cite{han2022card} introduces a classification and regression diffusion model to accurately capture the mean and the uncertainty of the prediction. 
DiffusionDet~\cite{chen2023diffusiondet} formulates object detection as a denoising diffusion process from noisy boxes to object boxes. 
Baranchuk \etal~\cite{baranchuk2021label} employ diffusion models for semantic segmentation with scarce labeled data. 
DiffusionInst~\cite{gu2022diffusioninst} depicts instances as instance-aware filters and casts instance segmentation as a denoising process from noise to filter. 
Jiang \etal~\cite{jiang2024se} introduce diffusion models to point cloud registration that operates on the rigid body transformation group. 
Recent research on optical flow and depth estimation~\cite{saxena2024surprising} shows the possibility of using diffusion models for dense vision tasks. 
While there have been attempts to employ diffusion models for scene flow estimation~\cite{liu2023difflow3d}, they mainly focus on refining an initial estimation. 
On the contrary, our goal is to construct a model to estimate the full scene flow vector field instead of a refinement plug-in module. 
To the best of our knowledge, we are the first to propose using diffusion models to estimate the full scene flow directly from two point clouds.

\section{Proposed Method}
\subsection{Preliminaries}
\label{sec: preliminaries}
\paragraph{Scene Flow Estimation.}
Given a source point cloud $\mathbf{P}_\mathrm{source} \in \mathbb{R}^{N_1 \times 3}$ and a target point cloud $\mathbf{P}_\mathrm{target} \in \mathbb{R}^{N_2 \times 3}$, where $N_1$ and $N_2$ are the number of points in the source and the target point cloud respectively, the objective is to estimate a scene flow vector field $\mathbf{V} \in \mathbb{R}^{N_1 \times 3}$ that maps each source point to the correct position in the target point cloud.

\paragraph{Diffusion Models.}
Inspired by non-equilibrium thermodynamics, diffusion models~\cite{ho2020denoising, song2021denoising} are a class of latent variable ($x_1, ..., x_T$) models of the form 
$p_{\theta}(x_{0}) = \int p_{\theta}(x_{0:T}) d x_{1:T}$, 
where the latent variables are of the same dimensionality as the input data $x_0$ (any dimensionality). 
The joint distribution $p_{\theta}(x_{0:T})$ is also called the \textit{reverse process}
\begin{equation} \label{eq: reverse}
\begin{array}{cc}
    p_{\theta}(x_{0:T}) = p(x_T)\prod_{t=1}^{T} p_{\theta}(x_{t-1}|x_{t}), & p_{\theta}(x_{t-1}|x_{t}) = \mathcal{N}(x_{t-1};\mu_\theta(x_t,t),\Sigma_\theta(x_t,t))
    . 
\end{array}
\end{equation}
The approximate posterior $q(x_{1:T}|x_{0})$ is called the \textit{forward process}, which is fixed to a Markov chain that gradually adds noise according to a predefined noise scheduler $\beta_{1:T}$
\begin{equation} \label{eq: diffusion}
\begin{array}{cc}
    q(x_{1:T}|x_{0}) = \prod_{t=1}^{T} q(x_{t}|x_{t-1}), & q(x_{t}|x_{t-1}) = \mathcal{N}(x_{t};\sqrt{1-\beta_{t}}x_{t-1},\beta_{t}\mathbf{I}).
\end{array}
\end{equation}
The training is performed by minimizing a variational bound on the negative log-likelihood 
\begin{equation} \label{eq: elbo}
\begin{split}
    \mathbb{E}_{q}[-\log p_\theta(x_{0})] &\leq \mathbb{E}_{q}[-\log \tfrac{p_\theta(x_{0:T})}{q(x_{1:T}|x_0)}] \\
    &= \mathbb{E}_{q}[D_\mathrm{KL}(q(x_{T}|x_{0})\|p(x_{T})) \\
    &+ \textstyle\sum_{t>1}D_{\mathrm{KL}}(q(x_{t-1}|x_{t}, x_{0})\|p_\theta(x_{t-1}|x_{t})) - \log p_\theta(x_0|x_1)],
\end{split}
\end{equation}
where $D_\mathrm{KL}$ denotes the Kullback–Leibler divergence.

\subsection{Scene Flow Estimation as Diffusion Process}
\label{sec: method}
We formulate the scene flow estimation task as a conditional diffusion process that is illustrated in Figure~\ref{Diffusion}. 
The \textit{forward process} starts from the ground truth scene flow vector field $\mathbf{V}_0$ and ends at pure Gaussian noise $\mathbf{V}_T$ by gradually adding Gaussian noise to the input data as in Eq.~\eqref{eq: diffusion}. 
Given that $\beta_t$ is small, $q(\mathbf{V}_{t}|\mathbf{V}_{t-1})$ in Eq.~\eqref{eq: diffusion} has a closed form~\cite{ho2020denoising}
\begin{equation}
    q(\mathbf{V}_{t}|\mathbf{V}_{0}) = \mathcal{N}(\mathbf{V}_{t};\sqrt{\bar{\alpha}_{t}}\mathbf{V}_{0},(1-\bar{\alpha}_{t})\mathbf{I}),
\end{equation}
where $\bar{\alpha}_{t}:=\prod_{s=1}^{t}(1-\beta_s)$.
The \textit{reverse process} predicts the ground truth $\mathbf{V}_{0}$ from the noisy input $\mathbf{V}_{t}$ conditioned on both the source point cloud $\mathbf{P}_\mathrm{source}$ and the target point cloud $\mathbf{P}_\mathrm{target}$, 
\begin{equation} \label{eq: reverse diffsf}
    p_\theta(\mathbf{V}_{t-1}|\mathbf{V}_{t}, \mathbf{P}_\mathrm{source}, \mathbf{P}_\mathrm{target}) = \mathcal{N}(\mathbf{V}_{t-1};\mu_\theta(\mathbf{V}_t, \mathbf{P}_\mathrm{source}, \mathbf{P}_\mathrm{target}),\textbf{I}).
\end{equation}
The forward process posterior is tractable when conditioned on $\mathbf{V}_{0}$,
\begin{equation} \label{eq: forward posterior}
    q(\mathbf{V}_{t-1}|\mathbf{V}_{t},\mathbf{V}_{0}) = \mathcal{N}(\mathbf{v}_{t-1};\tilde{\mu}_{t}(\mathbf{V}_{t},\mathbf{V}_{0}),\tilde{\beta}_{t}\mathbf{I}),
\end{equation}
where $\tilde{\mu}_{t}(\mathbf{V}_{t},\mathbf{V}_{0}) := \tfrac{\sqrt{\bar{\alpha}_{t-1}}\beta_{t}}{1-\bar{\alpha}_{t}}\mathbf{V}_{0}+\tfrac{\sqrt{\alpha_{t}}(1-\bar{\alpha}_{t-1})}{1-\bar{\alpha}_{t}}\mathbf{V}_{t}$, and $\tilde{\beta}_{t} := \tfrac{1-\bar{\alpha}_{t-1}}{1-\bar{\alpha}_{t}}\beta_{t}$. 
Minimizing the variational bound in Eq.~\eqref{eq: elbo} breaks down to minimizing the difference between $\tilde{\mu}_{t}(\mathbf{V}_{t},\mathbf{V}_{0})$ and $\mu_\theta(\mathbf{V}_t, \mathbf{P}_\mathrm{source}, \mathbf{P}_\mathrm{target})$. Since $\mathbf{V}_t$ is constructed from $\mathbf{V}_0$ by a predefined fixed noise scheduler $\beta_{1:T}$, the training objective is further equivalent to learning $\mathbf{V}_{0}$ by a neural network $f_\theta(\mathbf{V}_t, \mathbf{P}_\mathrm{source}, \mathbf{P}_\mathrm{target})$. The training loss can be written as 
\begin{equation} \label{eq: diffusion objective}
    \mathcal{L} = \|f_\theta(\mathbf{V}_t, \mathbf{P}_\mathrm{source}, \mathbf{P}_\mathrm{target}) - \mathbf{V}_0\|,
\end{equation}
where the neural network $f_\theta(\mathbf{V}_t, \mathbf{P}_\mathrm{source}, \mathbf{P}_\mathrm{target})$ takes the current noisy input $\mathbf{V}_t$, the source point cloud $\mathbf{P}_\mathrm{source}$, and the target point cloud $\mathbf{P}_\mathrm{target}$ as input and output $\hat{\mathbf{V}}_\mathrm{pred}$, which is an prediction of $\mathbf{V}_{0}$. The detailed architecture of $f_\theta$ is presented in section~\ref{sec: architecture}. 
The reverse process in Eq.~\eqref{eq: reverse diffsf} can be rewritten by replacing $\mu_{\theta}$ with $f_{\theta}$ as
\begin{equation} \label{eq: reverse process}
    p_\theta(\mathbf{V}_{t-1}|\mathbf{V}_t, \mathbf{P}_\mathrm{source}, \mathbf{P}_\mathrm{target}) = \mathcal{N}(\mathbf{V}_{t-1};\tilde{\mu}_{t}(\mathbf{V}_{t},f_\theta(\mathbf{V}_t, \mathbf{P}_\mathrm{source}, \mathbf{P}_\mathrm{target})),\textbf{I}).
\end{equation}

During \textit{inference}, starting from randomly sampled Gaussian noise $\mathbf{V}_T$, $\mathbf{V}_0$ is reconstructed with the model $f_\theta$ according to the reverse process in Eq.~\eqref{eq: reverse process}. 
The detailed training and sampling algorithms are given in Algorithm \ref{algo:training} and Algorithm \ref{algo:sampling}.

\noindent
\begin{minipage}{0.5\linewidth}
\begin{algorithm}[H]
  \caption{Training}
  \label{algo:training}
  \scriptsize
  \SetAlgoNlRelativeSize{-1}
  \Repeat{converged}{
    $\mathbf{V}_0 \sim q(\mathbf{V}_0)$, $\mathbf{\epsilon} \sim \mathcal{N}(\mathbf{0}, \mathbf{I})$\;
    $t \sim \text{Uniform}(\{1,...,T\})$\;
    $\mathbf{V}_t = \sqrt{\Bar{\alpha}_t}\mathbf{V}_0 + \sqrt{1-\Bar{\alpha}_t}\mathbf{\epsilon}$\;
    estimate $\hat{\mathbf{V}}_\mathrm{pred} = f_\theta(\mathbf{V}_t, \mathbf{P}_\mathrm{source}, \mathbf{P}_\mathrm{target})$\;
    optimize loss: $\mathcal{L}_t = \text{loss}(\hat{\mathbf{V}}_\mathrm{pred}, \mathbf{V}_0)$\;
  }
\end{algorithm}
\end{minipage}
\begin{minipage}{0.5\linewidth}
\begin{algorithm}[H]
  \caption{Sampling}
  \label{algo:sampling}
  \scriptsize
  \SetAlgoNlRelativeSize{-1}
  $\mathbf{V}_T \sim \mathcal{N}(\textbf{0},\textbf{I})$\;
  \For{$t = T, ..., 1$}{
    estimate $\hat{\mathbf{V}}_\mathrm{pred} = f_\theta(\mathbf{V}_t, \mathbf{P}_\mathrm{source}, \mathbf{P}_\mathrm{target})$\;
    if $t>1$: $\textbf{z} \sim \mathcal{N}(\textbf{0}, \textbf{I})$\;
    else: $\textbf{z} = \textbf{0}$\;
    $\mathbf{V}_{t-1} =  \Tilde{\mu}_{t}(\mathbf{V}_t,\hat{\mathbf{V}}_\mathrm{pred}) + \textbf{z}$\;
  }
  \Return{$\mathbf{V}_0$}\;
\end{algorithm}
\end{minipage}

\subsection{Architecture}
\label{sec: architecture}
\begin{figure}[t!]
    \centering
    \includegraphics[width=1.\linewidth]{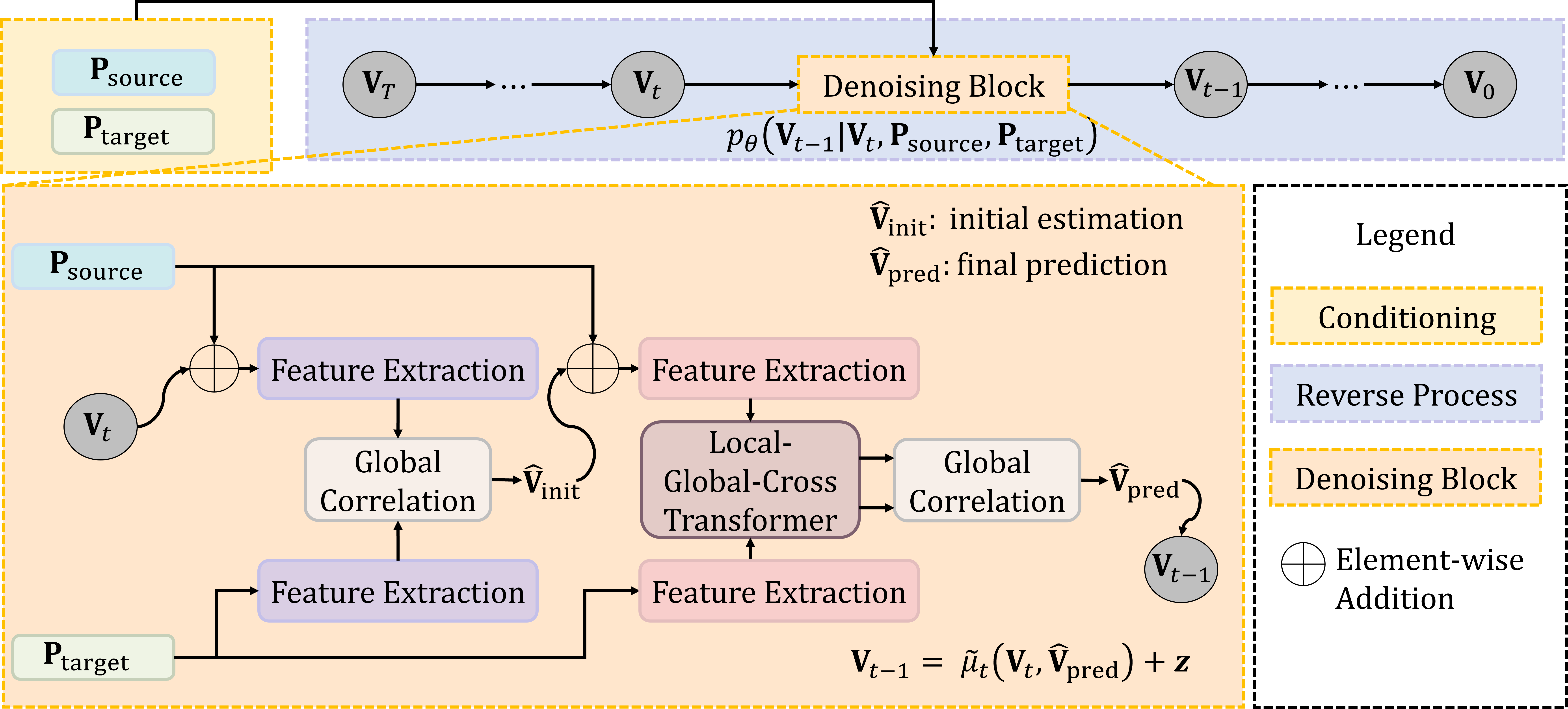}
    \caption{The reverse process with detailed denoising block for scene flow estimation. The denoising block takes the current noisy input $\mathbf{V}_t$, the source point cloud $\mathbf{P}_\mathrm{source}$, and the target point cloud $\mathbf{P}_\mathrm{target}$ as input. The output $\hat{\mathbf{V}}_\mathrm{pred}$ is the denoised scene flow prediction. Shared weights for the feature extraction are indicated in the same color.}
    \label{Framework}
\end{figure}

To train the diffusion process with Eq.~\eqref{eq: diffusion objective}, we need to design the neural network to predict $\mathbf{V}_0$, i.e. the ground truth scene flow vector field. 
The reverse process with the detailed architecture of $\hat{\mathbf{V}}_{\mathrm{pred}}=f_\theta(\mathbf{V}_t, \mathbf{P}_\mathrm{source}, \mathbf{P}_\mathrm{target})$ is given in Figure~\ref{Framework}. 
We take the state-of-the-art method GMSF~\cite{zhang2024gmsf} as our baseline. 
All the building blocks, Feature Extraction, Local-Global-Cross Transformer, and Global Correlation are the same as in GMSF~\cite{zhang2024gmsf}. 
We modify the model architecture of GMSF following the recent work~\cite{puy2020flot, gu2022rcp, kittenplon2021flowstep3d} of scene flow estimation by adding an initial estimation before the final prediction. 
More specifically, the source point cloud $\mathbf{P}_\mathrm{source} \in \mathbb{R}^{N_{1} \times 3}$ is first warped with $\mathbf{V}_{t} \in \mathbb{R}^{N_{1} \times 3}$. 
The warped source point cloud and the target point cloud are sent to the Feature Extraction block to expand the three-dimensional coordinate into higher-dimensional features for each point. 
Based on the similarities between point pairs in the warped source and the target point cloud, a Global Correlation is applied to compute an initial estimation $\hat{\mathbf{V}}_{\mathrm{init}} \in \mathbb{R}^{N_{1} \times 3}$. 
We then warp the source point cloud $\mathbf{P}_\mathrm{source} \in \mathbb{R}^{N_{1} \times 3}$ with the initial estimation $\hat{\mathbf{V}}_{\mathrm{init}} \in \mathbb{R}^{N_{1} \times 3}$. The same Feature Extraction block is applied on both the warped source point cloud and the target point cloud, but with different weights than the previous block. 
A Local-Global-Cross Transformer is then applied to the higher-dimensional features to get a more robust and reliable feature representation for each point. 
The output features are then sent into the Global Correlation block to get the final prediction $\hat{\mathbf{V}}_{\mathrm{pred}} \in \mathbb{R}^{N_{1} \times 3}$. 
The detailed architecture of Feature Extraction, Local-Global-Cross Transformer, and Global Correlation is given in the following paragraphs using the same notation as GMSF~\cite{zhang2024gmsf}.

\paragraph{Feature Extraction} The three-dimensional coordinate for each point is first projected into a higher feature dimension $\mathbf{x}_{i}^h \in \mathbb{R}^{1 \times d}$ by the off-the-shelf feature extraction backbone DGCNN~\cite{wang2019dynamic}. Each layer of the network can be written as
\begin{equation}
    \mathbf{x}_{i}^h = \max_{\mathbf{x}_{j} \in \mathcal{N}(i)} h(\mathbf{x}_i, \mathbf{x}_{j}-\mathbf{x}_{i}),
\end{equation}
where $i$ and $j$ denote the index of a single point in the point cloud. 
$\mathbf{x}_{j} \in \mathcal{N}(i)$ denotes the neighboring points of point $\mathbf{x}_{i}$ found by a $k$-nearest-neighbor (KNN) algorithm. The number of $k$ is set to 16. 
The point feature $\mathbf{x}_{i}$ and the edge feature $\mathbf{x}_{j}-\mathbf{x}_{i}$ are first concatenated together along the feature dimension and then passed through a neural network $h$. 
$h$ consists of a sequence of linear layer, batch normalization, and leaky ReLU layer. 
The output feature dimension $d$ is set to 128. 
The maximum value of the $k$ nearest neighbors is taken as the output. Multiple layers are stacked together to get the final feature representation $\mathbf{x}_{i}^h$. 

\paragraph{Local-Global-Cross Transformer} takes the output high-dimensional features $\mathbf{x}_{i}^{h} \in \mathbb{R}^{1 \times d}$ as input to learn more robust and reliable feature representations, 
\begin{equation} \label{eq: local transformer}
    \mathbf{x}_{i}^l = \textstyle\sum_{\mathbf{x}_j \in \mathcal{N}(i)}\gamma(\varphi_l(\mathbf{x}_i^h)-\psi_l(\mathbf{x}_j^h)+\delta) \odot (\alpha_l(\mathbf{x}_j^h)+\delta),
\end{equation}
\begin{equation} \label{eq: global transformer}
    \mathbf{x}_i^g = \textstyle\sum_{\mathbf{x}_j \in \mathcal{X}_1}\langle\varphi_g(\mathbf{x}_i^l), \psi_g(\mathbf{x}_j^l)\rangle \alpha_g(\mathbf{x}_j^l),
\end{equation}
\begin{equation} \label{eq: cross transformer}
    \mathbf{x}_{i}^c = \textstyle\sum_{\mathbf{x}_j \in \mathcal{X}_2}\langle\varphi_c(\mathbf{x}_i^g), \psi_c(\mathbf{x}_j^g)\rangle \alpha_c(\mathbf{x}_j^g),
\end{equation}
where local, global, and cross transformers are given in Eq.~\eqref{eq: local transformer}~\eqref{eq: global transformer}~\eqref{eq: cross transformer} respectively. 
$\varphi$, $\psi$, and $\alpha$ denote linear layers to generate the query, key, and value. 
The indices $\cdot_l$, $\cdot_g$, and $\cdot_c$ indicate local transformer, global transformer, and cross transformer, respectively. 
For the local transformer, 
$\gamma$ is a sequence of linear layer, ReLU, linear layer, and softmax. 
$\delta$ is the relative positional embedding that gives the information of the 3D coordinate distance between $\mathbf{x}_{i}$ and $\mathbf{x}_{j}$. 
$\odot$ denotes element-wise multiplication. 
The output $\mathbf{x}_{i}^{l}$ is further processed by a linear layer and a residual connection from the input before being sent to the global transformer. 
For the global and cross transformer, 
$\mathcal{X}_1 = \mathbf{P}_\mathrm{source} + (\mathbf{V}_{t}$ or $\hat{\mathbf{V}}_\mathrm{init}) \in \mathbb{R}^{N_{1} \times 3}$ and $\mathcal{X}_2 = \mathbf{P}_\mathrm{target} \in \mathbb{R}^{N_{2} \times 3}$ represent the warped source point cloud and the target point cloud, respectively. 
$\langle , \rangle$ denotes the scalar product. 
The output of the global and cross transformer is further processed by a linear layer, a layer normalization, and a residual connection from the input. 
A feedforward network with a multilayer perceptron and layer normalization is applied to the output of the cross transformer to aggregate information. 
To acquire more robust feature representations, the global-cross transformers are stacked and repeated multiple times ($14$ times in our experiment). 
For simplicity, we only give the equations for learning the features of $\mathcal{X}_1$. The features of $\mathcal{X}_2$ are computed by the same procedure. 
The output point features $\mathbf{x}_{i}^c$ and $\mathbf{x}_{j}^c$ for each point cloud are stacked together to form feature matrices $\mathbf{F}_1 \in \mathbb{R}^{N_{1} \times d}$ and $\mathbf{F}_2 \in \mathbb{R}^{N_{2} \times d}$. 

\paragraph{Global Correlation} predicts the scene flow vector solely based on two feature similarity matrices, cross feature similarity matrix $\mathbf{M}_\mathrm{cross} \in \mathbb{R}^{N_1 \times N_2}$ and self feature similarity matrix $\mathbf{M}_\mathrm{self} \in \mathbb{R}^{N_1 \times N_1}$. 
\begin{equation}
    \mathbf{M}_\mathrm{cross} = \text{softmax}(\mathbf{F}_1\mathbf{F}_2^T/ \sqrt{d}),
\end{equation}
\begin{equation}
    \mathbf{M}_\mathrm{self} = \text{softmax}(W_q(\mathbf{F}_1)W_k(\mathbf{F}_1)^T/ \sqrt{d}),
\end{equation}
where $W_q$ and $W_k$ are linear projections. 
$d$ is the feature dimensions. The softmax is taken over the second dimension of the matrices. 
The cross feature similarity matrix $\mathbf{M}_\mathrm{cross}\in \mathbb{R}^{N_1 \times N_2}$ encodes the feature similarities between all the points in the source point cloud $\mathbf{P}_\mathrm{source}$ and all the points in the target point cloud $\mathbf{P}_\mathrm{target}$. 
The self feature similarity matrix $\mathbf{M}_\mathrm{self}\in \mathbb{R}^{N_1 \times N_1}$ encodes the feature similarities between all points in the source point cloud $\mathbf{P}_\mathrm{source}$. 
The global correlation is performed by a matching process guided by the cross feature similarity matrix followed by a smoothing procedure guided by the self feature similarity matrix
\begin{equation}
   \hat{\mathbf{V}} = \mathbf{M}_\mathrm{self} (\mathbf{M}_\mathrm{cross} \mathbf{P}_\mathrm{target} - \mathbf{P}_\mathrm{source}).
\end{equation}
We follow GMSF~\cite{zhang2024gmsf} and employ a robust loss defined as
\begin{equation}
    \mathcal{L} = \textstyle\sum_{i}(\lVert\hat{\mathbf{V}}_{\mathrm{pred}}(i)-\mathbf{V}_{\mathrm{gt}}(i)\rVert_1+\epsilon)^q,
\end{equation}
where 
$\hat{\mathbf{V}}_{\mathrm{pred}}$ is the output prediction of the neural network, i.e. $f_\theta(\mathbf{V}_t, \mathbf{P}_\mathrm{source}, \mathbf{P}_\mathrm{target})$ in Eq.~\eqref{eq: diffusion objective}. 
$\mathbf{V}_{\mathrm{gt}}$ denotes the ground truth scene flow vector field i.e. $\mathbf{V}_0$ in Eq.~\eqref{eq: diffusion objective}. $i$ is the index of the points. $\epsilon$ is set to 0.01 and $q$ is set to 0.4. 

\section{Experiments}
\label{Experiments}
\subsection{Implementation Details}

We use the AdamW optimizer and a weight decay of $1 \times 10 ^ {-4}$. 
The initial learning rate is set to $4 \times 10 ^ {-4}$ for FlyingThings3D~\cite{mayer2016large} and $1 \times 10 ^ {-4}$ for Waymo-Open~\cite{sun2020scalability}.
We employ learning rate annealing by using the Pytorch OneCycleLR learning rate scheduler.
During training, we set $N_1$ and $N_2$ to $4096$, randomly sampled by furthest point sampling. The model is trained for $600$k iterations with a batch size of $24$.
During inference, we follow previous methods~\cite{zhang2024gmsf, liu2023difflow3d, cheng2023multi} and set $N_1$ and $N_2$ to $8192$ for a fair comparison. 
The number of diffusion steps is set to $20$ during training and $2$ during inference. 
The number of nearest neighbors $k$ in DGCNN  and Local Transformer is set to $16$. 
The number of global-cross transformer layers is set to $14$. 
The number of feature channels is set to $128$. 
Further implementation details are given in the supplemental document and the provided code.

\subsection{Evaluation Metrics}
We follow the most recent work in the field~\cite{zhang2024gmsf, liu2023difflow3d, cheng2023multi} and use established evaluation metrics for scene flow estimation. 
$\textrm{EPE}_{\textrm{3D}}$
measures the endpoint error between the prediction and the ground truth $\lVert\hat{\mathbf{V}}_\mathrm{pred}-\mathbf{V}_\mathrm{gt}\rVert_2$ averaged over all points.
$\textrm{ACC}_{\textrm{S}}$
measures the percentage of points with an endpoint error smaller than $5~\textrm{cm}$ or relative error less than $5\%$.
$\textrm{ACC}_{\textrm{R}}$
measures the percentage of points with an endpoint error smaller than $10~\textrm{cm}$ or relative error less than $10\%$.
$\textrm{Outliers}$
measures the percentage of points with an endpoint error larger than $30~\textrm{cm}$ or relative error larger than $10\%$.

\subsection{Datasets}
We follow the most recent work in the field~\cite{zhang2024gmsf, liu2023difflow3d, cheng2023multi} and test the proposed method on three established benchmarks for scene flow estimation. 

\textbf{FlyingThings3D}~\cite{mayer2016large} is a synthetic dataset consisting of 25000 scenes with ground truth annotations. 
We follow Liu \etal in FlowNet3D~\cite{liu2019flownet3d} and Gu \etal in HPLFlowNet~\cite{gu2019hplflownet} to preprocess the dataset and denote them as $\text{F3D}_\text{o}$, with occlusions, and $\text{F3D}_\text{s}$, without occlusions. 
The former consists of 20000 and 2000 scenes for training and testing, respectively. 
The latter consists of 19640 and 3824 scenes for training and testing, respectively. 

\textbf{KITTI Scene Flow}~\cite{menze2015object} is a real autonomous driving dataset with 200 scenes for training and 200 scenes for testing. 
Since the annotated data in KITTI is limited, the dataset is mainly used for evaluating the generalization ability of the models trained on FlyingThings3D.
Similar to the FlyingThings3D dataset, following Liu \etal in FlowNet3D~\cite{liu2019flownet3d} and Gu \etal in HPLFlowNet~\cite{gu2019hplflownet}, the KITTI dataset is preprocessed as $\text{KITTI}_\text{o}$, with occlusions, and $\text{KITTI}_\text{s}$, without occlusions.
The former consists of 150 scenes from the annotated training set. 
The latter consists of 142 scenes from the annotated training set. 

\textbf{Waymo-Open}~\cite{sun2020scalability} is a larger autonomous driving dataset with challenging scenes. 
The annotations are generated from corresponding tracked 3D objects to scale up the dataset for scene flow estimation by approximately $1000$ times compared to previous real-world scene flow estimation datasets. 
The dataset consists of 798 training sequences and 202 testing sequences.
Each sequence consists of around 200 scenes. 
Different preprocessing of the dataset exists~\cite{ding2022fh,jin2022deformation,jund2021scalable}, 
we follow the one employed in our baseline method~\cite{ding2022fh}. 

Note that Li et al.~\cite{li2021neural} preprocess datasets like Argoverse~\cite{chang2019argoverse} and nuScenes~\cite{caesar2020nuscenes} without providing corresponding training datasets. Therefore, these preprocessed datasets are suitable only for runtime optimization-based methods. 
In the absence of training data, several authors try to generate their own training datasets~\cite{li2023fast,jiang20243dsflabelling}, which means there is no standard protocol for evaluating learning-based methods on these datasets.  


\subsection{State-of-the-art Comparison}
We give state-of-the-art comparisons on multiple standard scene flow datasets. 
Table~\ref{F3Ds} and Table~\ref{F3Do} show the results on the $\text{F3D}_\text{s}$ and the $\text{F3D}_\text{o}$ datasets, with generalization results on the $\text{KITTI}_\text{s}$ and the $\text{KITTI}_\text{o}$ datasets. 
Table~\ref{Waymo} shows the results on the $\text{Waymo-Open}$ dataset. 
On the $\text{F3D}_\text{s}$ dataset, DiffSF shows an improvement (over the failure cases) of 31\% in $\textrm{EPE}_{\textrm{3D}}$, 44\% in $\textrm{ACC}_{\textrm{S}}$, 35\% in $\textrm{ACC}_{\textrm{R}}$, and 45\% in $\textrm{Outliers}$ compared to the current state-of-the-art method GMSF~\cite{zhang2024gmsf}. 
Similar improvement is also shown on the $\text{F3D}_\text{o}$ dataset with an improvement of 32\% in $\textrm{EPE}_{\textrm{3D}}$, 34\% in $\textrm{ACC}_{\textrm{S}}$, 24\% in $\textrm{ACC}_{\textrm{R}}$, and 38\% in $\textrm{Outliers}$, demonstrating DiffSF's ability to handle occlusions. 
The generalization abilities on the $\text{KITTI}_\text{s}$ and the $\text{KITTI}_\text{o}$ datasets are comparable to state of the art. All the four metrics show the best or second-best performances. On the Waymo-Open dataset, a steady improvement in both accuracy and robustness is achieved, demonstrating DiffSF's effectiveness on real-world data. 

\begin{table}[h!htb]
    \renewcommand{\arraystretch}{0.8}
    \setlength{\tabcolsep}{2pt}
    \begin{center}
    \small
    \caption{\textbf{State-of-the-art comparison on $\text{F3D}_\text{s}$ and $\text{KITTI}_\text{s}$.} The models are only trained on $\text{F3D}_\text{s}$ without occlusions. The number of time steps is set to 20 for training and 2 for inference. The bold and the underlined numbers represent the best and the second best performance respectively.}
    \label{F3Ds}
    \scalebox{1.0}{
    \begin{tabular}{l|cccc|cccc}
    \toprule
     Method & \multicolumn{4}{c}{$\text{F3D}_\text{s}$} & \multicolumn{4}{c}{$\text{KITTI}_\text{s}$}\\
     & $\text{EPE}_{\text{3D}}\downarrow$ & $\text{ACC}_{\text{S}}\uparrow$ & $\text{ACC}_{\text{R}}\uparrow$ & $\text{Outliers}\downarrow$ & $\text{EPE}_{\text{3D}}\downarrow$ & $\text{ACC}_{\text{S}}\uparrow$ & $\text{ACC}_{\text{R}}\uparrow$ & $\text{Outliers}\downarrow$ \\
    \midrule
    FlowNet3D~\cite{liu2019flownet3d}{\tiny CVPR'19} & 0.1136 & 41.25 & 77.06 & 60.16 & 0.1767 & 37.38 & 66.77 & 52.71 \\
    HPLFlowNet~\cite{gu2019hplflownet}{\tiny CVPR'19} & 0.0804 & 61.44 & 85.55 & 42.87 & 0.1169 & 47.83 & 77.76 & 41.03 \\
    PointPWC~\cite{wu2020pointpwc}{\tiny ECCV'20} & 0.0588 & 73.79 & 92.76 & 34.24 & 0.0694 & 72.81 & 88.84 & 26.48 \\
    FLOT~\cite{puy2020flot}{\tiny ECCV'20} & 0.0520 & 73.20 & 92.70 & 35.70 & 0.0560 & 75.50 & 90.80 & 24.20 \\
    Bi-PointFlow~\cite{cheng2022bi}{\tiny ECCV'22} & 0.0280 & 91.80 & 97.80 & 14.30 & 0.0300 & 92.00 & 96.00 & 14.10 \\
    3DFlow~\cite{wang2022matters}{\tiny ECCV'22} & 0.0281 & 92.90 & 98.17 & 14.58 & 0.0309 & 90.47 & 95.80 & 16.12 \\
    MSBRN~\cite{cheng2023multi}{\tiny ICCV'23} & 0.0150 & 97.30 & 99.20 & 5.60 & 0.0110 & 97.10 & 98.90 & 8.50 \\
    DifFlow3D~\cite{liu2023difflow3d}{\tiny CVPR'24} & 0.0140 & 97.76 & 99.33 & 4.79 & \textbf{0.0089} & \underline{98.13} & \underline{99.30} & \textbf{8.25} \\
    GMSF~\cite{zhang2024gmsf}{\tiny NIPS'23} & \underline{0.0090} & \underline{99.18} & \underline{99.69} & \underline{2.55} & 0.0215 & 96.22 & 98.25 & 9.84 \\
    \midrule
    \textbf{DiffSF(ours)} & \textbf{0.0062} & \textbf{99.54} & \textbf{99.80} & \textbf{1.41} & \underline{0.0098} & \textbf{98.59} & \textbf{99.44} & \underline{8.31} \\
    \bottomrule
    \end{tabular}}
    \vspace*{-4mm}
    \end{center}
\end{table}
\begin{table}[h!htb]
    \renewcommand{\arraystretch}{0.8}
    \setlength{\tabcolsep}{2pt}
    \begin{center}
    \small
    \caption{\textbf{State-of-the-art comparison on $\text{F3D}_\text{o}$ and $\text{KITTI}_\text{o}$.} The models are only trained on $\text{F3D}_\text{o}$ with occlusions. The number of time steps is set to 20 for training and 2 for inference.}
    \label{F3Do}
    \scalebox{1.0}{
    \begin{tabular}{l|cccc|cccc}
    \toprule
     Method & \multicolumn{4}{c}{$\text{F3D}_\text{o}$} & \multicolumn{4}{c}{$\text{KITTI}_\text{o}$}\\
     & $\text{EPE}_{\text{3D}}\downarrow$ & $\text{ACC}_{\text{S}}\uparrow$ & $\text{ACC}_{\text{R}}\uparrow$ & $\text{Outliers}\downarrow$ & $\text{EPE}_{\text{3D}}\downarrow$ & $\text{ACC}_{\text{S}}\uparrow$ & $\text{ACC}_{\text{R}}\uparrow$ & $\text{Outliers}\downarrow$ \\
    \midrule
    FlowNet3D~\cite{liu2019flownet3d}{\tiny CVPR'19} & 0.157 & 22.8 & 58.2 & 80.4 & 0.183 & 9.8 & 39.4 & 79.9 \\
    HPLFlowNet~\cite{gu2019hplflownet}{\tiny CVPR'19} & 0.168 & 26.2 & 57.4 & 81.2 & 0.343 & 10.3 & 38.6 & 81.4 \\
    PointPWC~\cite{wu2020pointpwc}{\tiny ECCV'20} & 0.155 & 41.6  & 69.9 & 63.8 & 0.118 & 40.3 & 75.7 & 49.6 \\
    FLOT~\cite{puy2020flot}{\tiny ECCV'20} & 0.153 & 39.6 & 66.0 & 66.2 & 0.130 & 27.8 & 66.7 & 52.9 \\
    Bi-PointFlow~\cite{cheng2022bi}{\tiny ECCV'22} & 0.073 & 79.1 & 89.6 & 27.4 & 0.065 & 76.9 & 90.6 & 26.4 \\
    3DFlow~\cite{wang2022matters}{\tiny ECCV'22} & 0.063 & 79.1 & 90.9 & 27.9 & 0.073 & 81.9 & 89.0 & 26.1 \\
    MSBRN~\cite{cheng2023multi}{\tiny ICCV'23} & 0.053 & 83.6 & 92.6 & 23.1 & 0.044 & 87.3 & 95.0 & 20.8 \\
    DifFlow3D~\cite{liu2023difflow3d}{\tiny CVPR'24} & 0.047 & 88.2 & 94.0 & 15.0 &  \textbf{0.029} & \textbf{95.9} & \textbf{97.5} & \textbf{10.8} \\
    GMSF~\cite{zhang2024gmsf}{\tiny NIPS'23} & \underline{0.022} & \underline{95.0} & \underline{97.5} & \underline{5.6} & 0.033 & 91.6 & 95.9 & 13.7 \\
    \midrule
    \textbf{DiffSF(ours)} &  \textbf{0.015} &  \textbf{96.7} &  \textbf{98.1} &  \textbf{3.5} & \textbf{0.029} & \underline{94.5} & \underline{97.00} & \underline{13.0} \\
    \bottomrule
    \end{tabular}}
    \vspace*{-6mm}
    \end{center}
\end{table}

\begin{table}[h!htb]
\renewcommand{\arraystretch}{0.8}
\setlength{\tabcolsep}{2pt}
    \centering
    \small
    \caption{\textbf{State-of-the-art comparison on Waymo-Open dataset.} The number of time steps is set to 20 for training and 2 for inference.}
    \label{Waymo}
    \medskip
    \scalebox{1.0}{
    \begin{tabular}{l|cccc} 
    \toprule
    Method & $\text{EPE}_{\text{3D}}\downarrow$ & $\text{ACC}_{\text{S}}\uparrow$ & $\text{ACC}_{\text{R}}\uparrow$ & $\text{Outliers}\downarrow$ \\
    \midrule
    FlowNet3D~\cite{liu2019flownet3d}{\tiny CVPR'19} & 0.225 & 23.0 & 48.6 & 77.9 \\
    PointPWC~\cite{wu2020pointpwc}{\tiny ECCV'20} & 0.307 & 10.3 & 23.1 & 78.6 \\
    FESTA~\cite{wang2021festa}{\tiny CVPR'21} & 0.223 & 24.5 & 27.2 & 76.5 \\
    FH-Net~\cite{ding2022fh}{\tiny ECCV'22} & 0.175 & 35.8 & 67.4 & 60.3 \\
    GMSF~\cite{zhang2024gmsf}{\tiny NIPS'23} & 0.083 & 74.7 & 85.1 & 43.5 \\
    \midrule
    \textbf{DiffSF(ours)} & \textbf{0.080} & \textbf{76.0} & \textbf{85.6} & \textbf{41.9} \\
    \bottomrule
    \end{tabular}}
\end{table}

\subsection{Uncertainty-error Correspondence}
One of the key advantages of our proposed method DiffSF compared to other approaches is that DiffSF can model uncertainty during inference, without being explicitly trained for this purpose. 
With uncertainty, we refer to the epistemic uncertainty, which reflects the confidence the model has in its predictions. 
In our case, we predict an uncertainty for the prediction of each point. 
We exploit the property of diffusion models to inject randomness into inherently deterministic tasks.
Without having to train multiple models, we predict multiple hypotheses using a single model with different initial randomly sampled noise. 

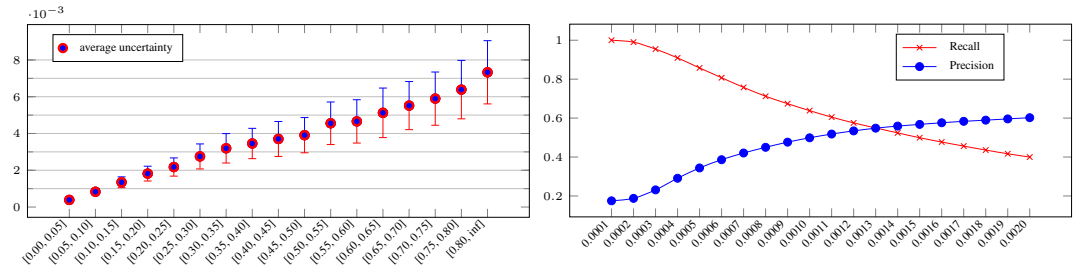
\begin{figure}[b!]
\begin{minipage}{0.5\textwidth}
\pgfplotstableread{
x y y-max y-min
{[0.00, 0.05]} 0.0003862025 0.00007097437628544867 0.00007097437628544867
{[0.05, 0.10]} 0.0008303055 0.0001703151036053896 0.0001703151036053896
{[0.10, 0.15]} 0.0013470065 0.00029484338592737913 0.00029484338592737913
{[0.15, 0.20]} 0.0018179301 0.0004024250898510218 0.0004024250898510218
{[0.20, 0.25]} 0.0021765835 0.0004949266090989113 0.0004949266090989113
{[0.25, 0.30]} 0.0027521912 0.00068046716041862965 0.00068046716041862965
{[0.30, 0.35]} 0.0031948113 0.0008005170151591301 0.0008005170151591301
{[0.35, 0.40]} 0.0034560033 0.000824122503399849 0.000824122503399849
{[0.40, 0.45]} 0.0037032177 0.0009500371292233467 0.0009500371292233467
{[0.45, 0.50]} 0.003909211 0.0009582859463989735 0.0009582859463989735
{[0.50, 0.55]} 0.004556134 0.0011576829478144646 0.0011576829478144646
{[0.55, 0.60]} 0.0046579326 0.0011779178865253925 0.0011779178865253925
{[0.60, 0.65]} 0.0051239394 0.0013473927974700928 0.0013473927974700928
{[0.65, 0.70]} 0.0055171438 0.00130817461758852 0.00130817461758852
{[0.70, 0.75]} 0.005896226 0.0014473450370132923 0.0014473450370132923
{[0.75, 0.80]} 0.006386147 0.0015884462743997574 0.0015884462743997574
{[0.80, inf]} 0.0073290495 0.0017202850431203842 0.0017202850431203842
}{\differanser}
\begin{tikzpicture}[scale=0.75] 
\begin{axis} [
width  = 1.5*\textwidth,
height = 5cm,
symbolic x coords={{[0.00, 0.05]},{[0.05, 0.10]},{[0.10, 0.15]},{[0.15, 0.20]},{[0.20, 0.25]},{[0.25, 0.30]},{[0.30, 0.35]},{[0.35, 0.40]},{[0.40, 0.45]},{[0.45, 0.50]},{[0.50, 0.55]},{[0.55, 0.60]},{[0.60, 0.65]},{[0.65, 0.70]},{[0.70, 0.75]},{[0.75, 0.80]},{[0.80, inf]}},
minor ytick={0.001,0.002,0.003,0.004,0.005,0.006,0.007,0.008},
yminorgrids,
xtick=data,
ticklabel style = {font=\tiny},
x tick label style={rotate=45,anchor=east},
legend style={at={(0.05,0.95)},anchor=north west,cells={anchor=west},column sep=1ex, font=\tiny}
]
\addplot+[blue, very thick, forget plot,only marks,forget plot] 
plot[very thick, error bars/.cd, y dir=plus, y explicit]
table[x=x,y=y,y error expr=\thisrow{y-max}] {\differanser};
\addplot+[red, very thick, only marks,xticklabels=\empty,forget plot] 
plot[very thick, error bars/.cd, y dir=minus, y explicit]
table[x=x,y=y,y error expr=\thisrow{y-min}] {\differanser};
\addplot[only marks,mark=*,mark options={fill=blue,draw=red,very thick}] 
table[x=x,y expr=\thisrow{y}] {\differanser};
\addlegendentry{average uncertainty}
\addplot[only marks,mark=-*,color=black] 
table[x=x,y expr=\thisrow{y}+\thisrow{y-max}] {\differanser};
\addplot[only marks,mark=-*,color=black] 
table[x=x,y expr=\thisrow{y}-\thisrow{y-min}] {\differanser};
\end{axis} 
\end{tikzpicture}
\end{minipage}
\begin{minipage}{0.5\textwidth}
\pgfplotstableread{
x y z 
{0.0001} 1.0000 0.1750
{0.0002} 0.9903 0.1876
{0.0003} 0.9544 0.2312
{0.0004} 0.9092 0.2909
{0.0005} 0.8577 0.3439
{0.0006} 0.8074 0.3862
{0.0007} 0.7577 0.4206
{0.0008} 0.7121 0.4499
{0.0009} 0.6740 0.4764
{0.0010} 0.6383 0.4987
{0.0011} 0.6050 0.5180
{0.0012} 0.5753 0.5342
{0.0013} 0.5494 0.5482
{0.0014} 0.5240 0.5589
{0.0015} 0.4991 0.5676
{0.0016} 0.4770 0.5760
{0.0017} 0.4562 0.5834
{0.0018} 0.4357 0.5894
{0.0019} 0.4168 0.5958
{0.0020} 0.3997 0.6021
}{\differanser}
\begin{tikzpicture}[scale=0.75] 
\begin{axis} [
width  = 1.5*\textwidth,
height = 5cm,
symbolic x coords={{0.0001},{0.0002},{0.0003},{0.0004},{0.0005},{0.0006},{0.0007},{0.0008},{0.0009},{0.0010},{0.0011},{0.0012},{0.0013},{0.0014},{0.0015},{0.0016},{0.0017},{0.0018},{0.0019},{0.0020}},
minor ytick={0.001,0.002,0.003,0.004,0.005,0.006,0.007,0.008},
yminorgrids,
xtick=data,
ticklabel style = {font=\tiny},
x tick label style={rotate=45,anchor=east},
legend style={at={(0.65,0.95)},anchor=north west,cells={anchor=west},column
sep=1ex, font=\tiny}
]
\addplot[smooth,color=red,mark=x] 
table[x=x,y expr=\thisrow{y}] {\differanser};
\addlegendentry{Recall}
\addplot[smooth,mark=*,blue] 
table[x=x,y expr=\thisrow{z}] {\differanser};
\addlegendentry{Precision}
\end{axis} 
\end{tikzpicture}
\end{minipage}
\caption{Analysis of uncertainty estimation on $\text{F3D}_\text{o}$ dataset. \textbf{Left}: Uncertainty-error correspondences. The horizontal axis is an interval of EPE. The vertical axis is the estimated uncertainty averaged over all the points that fall in the interval and the indication of the scaled uncertainty standard deviation. \textbf{Right}: Recall (red) and precision curve (blue) of outliers prediction. The horizontal axis is the threshold of the estimated uncertainty to determine the outliers.}
\label{Uncertainty}
\end{figure}

Figure~\ref{Uncertainty} shows that the standard deviation of $20$ hypotheses for each point gives a reliable uncertainty estimation, which correlates very well with the inaccuracy of the prediction. 
Figure~\ref{Uncertainty} (left) shows the relationship between the EPE and the standard deviation of the predictions averaged over the $\text{F3D}_\text{o}$ dataset.
There is an almost linear correlation of the predicted uncertainty with the EPE underlining the usefulness of our uncertainty measure. 
Figure~\ref{Uncertainty} (right) shows the recall and precision of the outlier prediction by the uncertainty. 
An outlier is defined as a point that has an EPE larger than 0.30 meters. 
The horizontal axis is the threshold applied to the uncertainty to determine the outliers. 
The recall is defined as the number of correctly retrieved outliers divided by the number of all the outliers. 
The precision is defined as the number of correctly retrieved outliers divided by the number of all the retrieved outliers. 
The precision-recall break-even point obtains around 55\% of recall and 55\% of precision. 

\begin{figure}[h!]
    \centering
    \includegraphics[width=1.\linewidth]{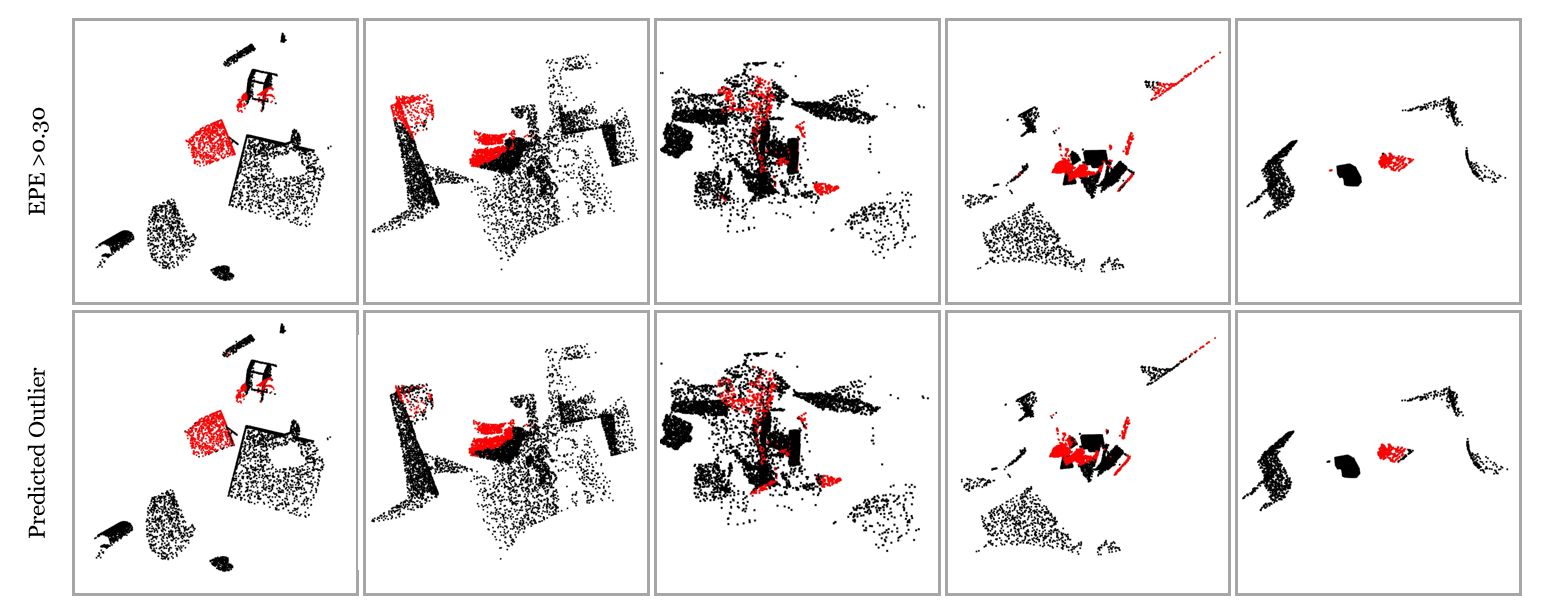}
    \caption{Visualization of outlier prediction on $\text{F3D}_\text{o}$ dataset. \textbf{Black}: Accurate prediction. \textbf{Red}: Outliers. Top row: Outliers defined as EPE > 0.30. Bottom row: Outliers predicted by Uncertainty.}
    \label{visualization}
\end{figure}

Figure~\ref{visualization} shows visual examples that compare our outlier prediction with the actual outliers.
The first row marks the scene flow estimation outliers with an EPE larger than 0.30 meters in red. 
The second row marks the outliers predicted by the uncertainty estimation in red.  
In summary, while every learned scene flow prediction model inevitably makes mistakes, our novel formulation of the task as a diffusion process not only produces state-of-the-art results but also allows for an accurate prediction of these errors.
Moreover, our analysis shows that downstream tasks can select a threshold according to its desired precision and recall, therefore, mitigating potential negative effects that uncertain predictions might produce.

\subsection{Ablation Study}
We investigate several key design choices of the proposed method. 
For the denoising model architecture, we investigate how the number of global-cross transformer layers and the number of feature channels affect the results.
For the diffusion process, we investigate the influence of the number of time steps for training and sampling.

\noindent\textbf{Model Architecture.}
To evaluate different architectural choices we select a diffusion model with five denoising blocks during training and one denoising step during testing with the DDIM~\cite{song2021denoising} sampling strategy. 
Table~\ref{layers} shows the influence of the number of global-cross transformer layers on the results. 
The experiments show that the best performance is achieved at the number of $14$ layers.
Table~\ref{channels} shows the influence of the number of feature channels on the results. 
The experiments show that a smaller number of feature channels results in worse performance. 
The best performance is achieved at $128$ feature channels.

\noindent\textbf{Number of Time Steps.}
We set the number of global-cross transformer layers to $14$ and the number of feature channels to $128$.
We investigate the influence of different number of time steps during training and sampling on the results. 
The number of time steps investigated is $5$, $20$, and $100$ for training and $1$, $2$, $5$, and $20$ for sampling. 
The fast sampling is done by DDIM~\cite{song2021denoising} instead of DDPM~\cite{ho2020denoising} sampling.
Table~\ref{Ablation study on the number of time steps for training and sampling} shows the results on the $\text{F3D}_\text{o}$ dataset, where $a@b$ denotes using $b$ training steps and $a$ sampling steps. 
While the results are very stable across a wide range of values, the best performance is achieved at $2@20$ time steps. 
We hypothesize that compared to the standard setting of image generation, the lower dimensionality and variance of the scene flow data results in a smaller number of required time steps. 
For the number of time steps during inference, DDIM sampling works well with the best performance achieved at $2$ steps. 

\begin{table}[h!htb]
    \renewcommand{\arraystretch}{0.8}
    \setlength{\tabcolsep}{2pt}
    \begin{center}
    \small
    \caption{Ablation study on the number of global-cross transformer layers on $\text{F3D}_\text{o}$. The number of feature channels is set to $128$. The number of time steps is set to $5$ for training and $1$ for inference.}
    \label{layers}
    \scalebox{1.0}{
    \begin{tabular}{c|cccc|cccc}
    \toprule
    Layers & $\text{EPE}_{\text{3D}}\downarrow$ & $\text{ACC}_{\text{S}}\uparrow$ & $\text{ACC}_{\text{R}}\uparrow$ & $\text{Outliers}\downarrow$ & $\text{EPE}_{\text{3D}}\downarrow$ & $\text{ACC}_{\text{S}}\uparrow$ & $\text{ACC}_{\text{R}}\uparrow$ & $\text{Outliers}\downarrow$ \\
    & \multicolumn{4}{c}{\text{all}} & \multicolumn{4}{c}{\text{non-occ}} \\
    \midrule
    8 & 0.0439 & 91.6 & 94.8 & 7.9 & 0.0205 & 95.2 & 97.5 & 5.1  \\
    10 & 0.0413 & 92.6 & 95.1 & 7.1 & 0.0189 & 95.8 & 97.6 & 4.5  \\
    12 & 0.0381 & 93.0 & 95.5 & 6.4 & 0.0168 & 96.1 & 97.8 & 3.9 \\
    14 & \textbf{0.0361} & \textbf{93.7} &  \textbf{95.7} & \textbf{5.9} & \textbf{0.0153} & \textbf{96.5} & \textbf{98.0} & \textbf{3.5} \\
    16 & 0.0383 & 93.0 & 95.5 & 6.5 & 0.0168 & 96.1 & 97.8 & 4.0 \\
    \bottomrule
    \end{tabular}}
    \end{center}
\end{table}

\begin{table}[h!htb]
    \renewcommand{\arraystretch}{0.8}
    \setlength{\tabcolsep}{2pt}
    \begin{center}
    \small
    \caption{Ablation study on the number of feature channels on $\text{F3D}_\text{o}$. The number of global-cross transformer layers is set to $14$. The number of time steps is set to $5$ for training and $1$ for inference.}
    \label{channels}
    \scalebox{1.0}{
    \begin{tabular}{c|cccc|cccc}
    \toprule
    Channels & $\text{EPE}_{\text{3D}}\downarrow$ & $\text{ACC}_{\text{S}}\uparrow$ & $\text{ACC}_{\text{R}}\uparrow$ & $\text{Outliers}\downarrow$ & $\text{EPE}_{\text{3D}}\downarrow$ & $\text{ACC}_{\text{S}}\uparrow$ & $\text{ACC}_{\text{R}}\uparrow$ & $\text{Outliers}\downarrow$ \\
    & \multicolumn{4}{c}{\text{all}} & \multicolumn{4}{c}{\text{non-occ}} \\
    \midrule
    32 & 0.0612 & 88.2 & 92.9 & 11.7 & 0.0299 & 92.9 & 96.3 & 8.2 \\
    64 & 0.0431 & 92.3 & 95.0 & 7.4 & 0.0199 & 95.7 & 97.5 & 4.7 \\
    128 & \textbf{0.0361} & \textbf{93.7} &  \textbf{95.7} & \textbf{5.9} & \textbf{0.0153} & \textbf{96.5} & \textbf{98.0} & \textbf{3.5} \\
    \bottomrule
    \end{tabular}}
    \end{center}
\end{table}

\begin{table}[h!htb]
    \renewcommand{\arraystretch}{0.8}
    \setlength{\tabcolsep}{2pt}
    \begin{center}
    \small
    \caption{Ablation study on the number time steps for training and sampling on $\text{F3D}_\text{o}$. The number of global-cross transformer layers is set to $14$. The number of feature channels is set to $128$. $a@b$ denotes an inference of $b$ training steps and $a$ sampling steps.}
    \label{Ablation study on the number of time steps for training and sampling}
    \scalebox{1.0}{
    \begin{tabular}{l|cccc|cccc}
    \toprule
    Steps & $\text{EPE}_{\text{3D}}(\textrm{cm})\downarrow$ & $\text{ACC}_{\text{S}}\uparrow$ & $\text{ACC}_{\text{R}}\uparrow$ & $\text{Outliers}\downarrow$ & $\text{EPE}_{\text{3D}}(\textrm{cm})\downarrow$ & $\text{ACC}_{\text{S}}\uparrow$ & $\text{ACC}_{\text{R}}\uparrow$ & $\text{Outliers}\downarrow$ \\
    & \multicolumn{4}{c}{\text{all}} & \multicolumn{4}{c}{\text{non-occ}} \\
    \midrule
    1@5 & 3.608 & 93.701 &  95.732 & 5.904 & 1.527 & 96.549 & 97.973 & 3.527 \\
    2@5 & 3.590 & 93.718 & 95.727 & 5.910 & 1.518 & 96.558 & 97.957 & 3.544 \\
    5@5 & 3.592 & 93.716 & 95.720 & 5.911 & 1.521 & 96.556 & 97.953 & 3.545 \\
    1@20 & 3.588 & \underline{93.870} & 95.912 & 5.798 & 1.504 & \underline{96.731} & 98.080 & 3.520 \\
    2@20 & \textbf{3.576} & \textbf{93.871} & \textbf{95.919} & \underline{5.791} & \textbf{1.491} & \textbf{96.736} & \textbf{98.083} & 3.511 \\
    5@20 & 3.580 & 93.865 & \underline{95.917} & \underline{5.791} & 1.492 & 96.730 & \textbf{98.083} & \textbf{3.507} \\
    20@20 & \underline{3.579} & 93.865 & 95.915 & \textbf{5.789} & \textbf{1.491} & \underline{96.731} & 98.082 & \underline{3.508} \\
    1@100 & 3.678 & 93.503 & 95.665 & 6.016 & 1.587 & 96.376 & 97.844 & 3.689 \\
    2@100 & 3.663 & 93.545 & 95.662 & 6.010 & 1.579 & 96.398 & 97.838 & 3.697 \\
    5@100 & 3.668 & 93.546 & 95.663 & 6.010 & 1.583 & 96.400 & 97.842 & 3.695 \\
    20@100 & 3.670 & 93.545 & 95.663 & 6.015 & 1.584 & 96.396 & 97.843 & 3.700 \\
    \bottomrule
    \end{tabular}}
    \end{center}
\end{table}

\noindent\textbf{Ablation study compare to baseline GMSF.}
To show the improvement of our method compared to the baseline GMSF~\cite{zhang2024gmsf}, we provide an additional ablation study on $\text{F3D}_\text{o}$. 
Since the original paper GMSF has a different training setting as our proposed DiffSF, for a fair comparison we retrain the GMSF baseline with our training setting. The result is given in Table~\ref{contributation} (first line).
The check in the two columns denotes the implementation of improved architecture and diffusion process, respectively. 
The results clearly show that the proposed method DiffSF achieves superior performance than GMSF. 
Both the improvement of the architecture and the introduction of the diffusion process contribute to the superior performance. 
The improved percentage (for the introduction of the diffusion process) over the failure case is marked in the table. 
The results show that the proposed method has a moderate improvement in the accuracy metric $\text{EPE}_{\text{3D}}$ and a huge improvement (more than $10\%$) in the robustness metrics $\text{ACC}_{\text{S}}$, $\text{ACC}_{\text{R}}$, and $\text{Outliers}$. 
Besides the better performance, the proposed method can also provide a per-prediction uncertainty. 
\begin{table}[h]
\renewcommand{\arraystretch}{0.8}
\setlength{\tabcolsep}{1pt}
    \centering
    \small
    \caption{Ablation Study compare to baseline GMSF on $\text{F3D}_\text{o}$.}
    \vspace{-2mm}
    \label{contributation}
    \medskip
    \scalebox{0.85}{
    \begin{tabular}{c|c|llll|llll} 
    \toprule
     improved & diffusion & \multicolumn{4}{c}{$\text{F3D}_\text{o}$-all} & \multicolumn{4}{c}{$\text{F3D}_\text{o}$-nonoccluded}\\
     architecture & process & $\text{EPE}_{\text{3D}}\downarrow$ & $\text{ACC}_{\text{S}}\uparrow$ & $\text{ACC}_{\text{R}}\uparrow$ & $\text{Outliers}\downarrow$ & $\text{EPE}_{\text{3D}}\downarrow$ & $\text{ACC}_{\text{S}}\uparrow$ & $\text{ACC}_{\text{R}}\uparrow$ & $\text{Outliers}\downarrow$ \\
    \midrule
     &  & 0.039 & 92.9 & 95.4 & 6.7 & 0.017 & 96.0 & 97.8 & 4.2 \\
     & \checkmark & 0.061 & 84.8 & 92.3 & 16.7 & 0.037 & 88.9 & 95.3 & 13.9 \\
    \checkmark &  & 0.037 & 93.2 & 95.4 & 6.5 & 0.016 & 96.2 & 97.7 & 4.1 \\
    \checkmark & \checkmark & 0.036\textcolor{blue}{(-2.7\%)} & 93.9\textcolor{blue}{(+10.3\%)} & 95.9\textcolor{blue}{(+10.9\%)} & 5.8\textcolor{blue}{(-10.8\%)} & 0.015\textcolor{blue}{(-6.3\%)} & 96.7\textcolor{blue}{(+13.2\%)} & 98.1\textcolor{blue}{(+17.4\%)} & 3.5\textcolor{blue}{(-14.6\%)} \\
    \bottomrule
    \end{tabular}}
    \vspace{-2mm}
\end{table}

\section{Conclusions}
\label{Conclusions}
We propose to estimate scene flow from point clouds using diffusion models in combination with transformers. 
Our novel approach provides significant improvements over the state-of-the-art in terms of both accuracy and robustness. 
Extensive experiments on multiple scene flow estimation benchmarks demonstrate the ability of DiffSF to handle both occlusions and real-world data. 
Furthermore, we propose to estimate uncertainty based on the randomness inherent in the diffusion process, which helps to indicate reliability for safety-critical downstream tasks. 
The proposed uncertainty estimation will enable mechanisms to mitigate the negative effects of potential failures. 

\parag{Limitations.}
The training process of the diffusion models relies on annotated scene flow ground truth which is not easy to obtain for real-world data. 
Incorporating self-supervised training methods to leverage unannotated data might further improve our approach in the future.
Furthermore, the transformer-based architecture and the global matching process limit the maximum number of points, and further research is required for peforming matching at scale. 


\parag{Acknowledgements.} 
This work was partly supported by the Wallenberg Artificial Intelligence, Autonomous Systems and Software Program (WASP), funded by Knut and Alice Wallenberg Foundation, and the Swedish Research Council grant 2022-04266;
and by the strategic research environment ELLIIT funded by the Swedish government.
The computational resources were provided by the National Academic Infrastructure for Supercomputing in Sweden (NAISS) at C3SE partially funded by the Swedish Research Council grant 2022-06725, and by the Berzelius resource, provided by the Knut and Alice Wallenberg Foundation at the National Supercomputer Centre.

{\small
\bibliographystyle{ieee_fullname}
\bibliography{diffsf}
}



\end{document}